\documentclass[10pt, conference, compsocconf]{IEEEtran}
\usepackage[utf8]{inputenc}
\usepackage{amssymb}
\usepackage{graphicx}
\usepackage[ruled, noend]{algorithm2e}
\usepackage{comment}
\usepackage{hyperref}
\usepackage{url}
\usepackage{multirow}
\usepackage{amsmath}
\usepackage{xcolor,colortbl}
\usepackage{float}
\usepackage{capt-of}

\def\B#1{\mathbf{#1}}

\definecolor{Gray}{gray}{0.92}
\newcolumntype{u}{>{\columncolor{Gray}}c}

\begin{document}
\title{Social-sparsity brain decoders: faster spatial sparsity}
\author{
\IEEEauthorblockN{
Ga\"el VAROQUAUX\IEEEauthorrefmark{1},
Matthieu KOWALSKI\IEEEauthorrefmark{1}\IEEEauthorrefmark{2},
Bertrand THIRION\IEEEauthorrefmark{1},
}
\IEEEauthorblockA{\IEEEauthorrefmark{1}INRIA Parietal,
Neurospin, b\^at 145, CEA Saclay, 91191 Gif sur Yvette, France\\
firstname.lastname@inria.fr
}
\IEEEauthorblockA{\IEEEauthorrefmark{2}L2S, Univ Paris-Sud -- CNRS -- CentraleSupelec
}}

\maketitle              

\begin{abstract}
Spatially-sparse predictors are good models for brain decoding: they give
accurate predictions and their weight maps are interpretable as they
focus on a small number of regions. However, the state of the art, based
on total variation or graph-net, is computationally costly. 
Here we introduce sparsity in the local neighborhood of each voxel
with social-sparsity, a structured shrinkage operator.
We find that, on brain imaging
classification problems, social-sparsity performs almost as well as
total-variation models and better than graph-net, for a fraction of the
computational cost. It also very clearly outlines predictive regions.
We give details of the model and the algorithm.
\end{abstract}

\begin{IEEEkeywords}
brain decoding, sparsity, spatial regularization
\end{IEEEkeywords}

\section{Introduction: Spatial sparsity in decoding}

Machine learning can predict behavior or
phenotype from brain images. Across subjects, it can give indications on a pathology or its
progression, for instance capturing atrophy patterns of Alzheimer-related
cognitive decline from anatomical imaging \cite{fan2008spatial}. 
To study brain function, it has been used extensively to predict some
cognitive parameters associated with the presented stimuli from
functional brain images such as functional Magnetic Resonance Imaging
(fMRI) \cite{haxby2001}.
In these
applications, the number of samples is small (hundreds or less), while
the number of features is typical the number of voxels in the brain,
50\,000 or more. The estimation is ill-posed, as there are much more
parameters to estimate than the number of samples, which calls for
regularization.

In brain decoding, good prediction is important, but also retrieving and
understanding the aspects of brain images that drive this prediction.
Linear models, as linear support vector machines (SVM), are often used
because they work well in the low-sample regime. An additional benefit of
these models is that their weights form brain maps
\cite{mouraomiranda2005}. However, suitable regularization is necessary
for these weights to retrieve well the important regions
\cite{varoquaux2012small,gramfort2013}. Sparse penalties select voxels
\cite{yamashita2008,carroll2009}, but only a subset of the important ones
\cite{varoquaux2012small}. Spatial and sparse penalties, using total
variation (TV) \cite{michel2011} or graph-net
\cite{grosenick2013interpretable}, help the decoder capture full regions
\cite{gramfort2013}.
TV and its variants are state-of-the-art regularizers for brain images.
In addition to decoding
\cite{michel2011,baldassarre2012,gramfort2013}, they have been used with
great success for applications as diverse as MR image reconstruction
\cite{huang2011efficient}, super-resolution \cite{tourbier2015efficient},
prediction from anatomical images \cite{dubois2014predictive}, and even
regularization of spatial registration \cite{fiot2014longitudinal}. The
main drawback of spatial sparsity as in TV and related penalties is its
computational cost.

Here we introduce spatial sparsity based on ``social sparsity'', a
relaxed structured penalty with a simple closed-form shrinkage operator.
We first detail the mathematical underpinnings of the model and then
perform decoding experiments that show that the model is almost as
accurate in prediction as the TV-based state of the art and much faster.

\paragraph*{Notations} Vectors are written in bold: $\B{x}$. Indexing
vectors is written as sub-scripts: $\B{x}_i$. In an iterative scheme, the
iteration number is a sub-script in parenthesis: $\B{x}_{(k)}$.

\section{Methods: Introducing social sparsity}

\subsection{Spatial penalties for neuroimaging}

A sparsity assumption on the decoders maps has been identified early as a
means to select voxels relevant for decoding
\cite{yamashita2008,carroll2009}. Sparsity has been used with great
success in statistics, signal processing, and machine learning. Indeed,
it combines good properties for prediction and denoising when the
ground truth is actually sparse \cite{hastie2015sparsity}. Sparse models can
be estimated by adding to the data-fit term an $\ell_1$ penalty for
sparsity, as in the famed Lasso. The Iterative Shrinkage/Thresholding Algorithm 
(ISTA) is a common algorithm to solve the corresponding minimization
problem \cite{daubechies2004iterative}. It alternates a gradient-descent
step for the data-fit energy and, for sparsity,
a \emph{soft-thresholding}: a proximal
operator which corresponds to the Euclidean projection on some $\ell_1$
ball:%
\begin{equation}
\ell_1 \text{ prox of } \B{w} \text{, scale }\lambda:
\quad\forall i
\quad\B{w}_i \leftarrow
\B{w}_i \, \biggl(1 -
\frac{\lambda}{|\B{w}_i|}\biggr)^{\!+}
\end{equation}
where the operation is applied element-wise --for every coordinate $\B{w}_i$
of $\B{w}$-- and $(.)^{+}$ is the positive part.
This operation is reminiscent of element-wise thresholding: it sets to
zero all the entries of $\B{w}$ that are smaller than $\lambda$ in
absolute value, and shrinks by $\lambda$ the remaining.

However, on brain images, simple sparse models as with the $\ell_1$
penalty select a subset of the important voxels
\cite{varoquaux2012small,rish2012,rondina2014scors}. Indeed as the
information in a voxel is very correlated to its neighbors, sparsity
focuses on one representative in a local neighborhood
\cite{varoquaux2012small}. Improved penalties add a spatial term to
couple neighbors and create structured sparsity. Total-variation (TV)
imposes sparsity in the image gradients \cite{michel2011}. Coupled with
sparsity in a TV-$\ell_1$ penalty it recovers very well the predictive
brain regions \cite{baldassarre2012,gramfort2013}. It is the
state of the art for brain decoding in terms of predictive power and 
of interpretability of the decoder maps.
The main drawback of total variation is that it leads to very
slow solvers. Indeed, it cannot be formulated in terms of a
thresholding-like operator. Its proximal operator is computational
costly as it couples all the voxels. Another spatial penalty, related but
faster, Graph-net, imposes smoothness, rather than sparsity, on the image
gradients \cite{grosenick2013interpretable}. Because it is a
differentiable penalty, it leads to faster optimization, though it
imposes less spatial structure.

The challenges of solving TV and related penalties arise from their very
strength: they impose sparsity not on the voxels of the image, but on a
representation capturing differences between voxels. This
concept is an instance of \emph{analysis sparsity}, which leads to other
penalties successful for predicting from brain images
\cite{eickenberg2015total}. Overlapping group sparsity is another
analysis penalty which has been heavily used in image processing
\cite{mairal2014sparsemodelling}: rather than putting voxels to zeros, it
penalizes a full local neighborhood, using a penalty on blocks. As a
voxel is in several neighborhoods, these blocks are overlapping. This
overlap leads to an optimization problem that has the same
structure --and same cost-- as that of TV-$\ell_1$.

All these estimators, and related optimization problems, can be solved
with an ISTA algorithm, or its accelerated variant the FISTA. In fact,
these approaches are the best option in the case of brain decoding
\cite{dohmatob2014benchmarking} although the TV-$\ell_1$ proximal
operator involves a costly sub-iteration.
In the case of non-overlapping blocks, group sparsity is simply 
$\ell_{21}$ penalty applied on each group, as in the famed group-lasso.
The proximal operator is
closed-form and performs a soft-thresholding on each group based on its
Euclidean norm:
\begin{equation}
\parbox{.18\linewidth}{$\ell_{21}$ prox\\
on group $\mathcal{G}$}:
\quad\forall i \in \mathcal{G}
\quad\B{w}_i \leftarrow
\B{w}_i \, \biggl(1 -
\frac{\lambda}{\sqrt{\sum_{j \in \mathcal{G}} \B{w}_j^2}}\biggr)^{\!+}
\end{equation}
for a coordinate $\B{w}_i$ in the group $\mathcal{G}$. As with 
soft-thresholding for the $\ell_1$ penalty, the operation is applied 
element-wise for all the groups and leads to fast optimizations.

\subsection{Fast spatial structure with social sparsity}

There is a large gap in computational cost between sparsity imposed on
separate items, coordinates or groups, and coupling the sparsity
of a voxel to that of its neighbor, as in overlapping group sparsity or
penalties related to TV. We introduce social sparsity
\cite{kowalski2013social} which is a tradeoff between the two scenarios.
It applies a soft-thresholding similar to group-lasso, using the norm of
a local neighborhood, but modifies only the coordinate $\B{w}_i$ at the
center of this neighborhood. In a sense, social sparsity ``forgets''
overlaps across neighborhoods. This makes solvers much faster, as
we show in the following.

\paragraph{The social-sparsity operator}
We focus here on the most popular "social-sparse" operator: the windowed group-lasso. For each element $\B{w}_i$, we associate a neighborhood $\mathcal{N}(i)$ of coordinates. The shrinkage operator reads 
\begin{equation}
\parbox{.16\linewidth}{$S(\B{w}, \lambda)$ social shrinkage}:
\quad\forall i
\quad\B{w}_i \leftarrow
\B{w}_i \, \biggl(1 -
\frac{\lambda}{\sqrt{\sum_{j \in \mathcal{N}(i)} \alpha_j^{i} \B{w}_j^2}}\biggr)^{\!+}
\label{eq:social_shrinkage}
\end{equation}
where the $\alpha_j^{i}$ are weights representing the shape of the
neighborhood (the simplest choice being the rectangular window:
all $\alpha_j^{i}$ are equal). 

An interesting interpretation of this operator, is that it can be seen as
a fast approximation of the group-lasso with overlaps as presented
in~\cite{jacob2009group}. Indeed, it can be shown that the social-sparse
operator is equivalent to applying a regular group-lasso operator in a high
dimensional space where all the variables are duplicated in order to form
independent groups (there are as many groups as variables). Then the result is projected back in the original space following an oblique projection~\cite{kowalski2013social}. In practice, similar performances are obtained 
without the cost 
encountered by the group-lasso with overlaps.

\paragraph{Choice of the neighborhood}
Coupling neighboring voxels is crucial, as demonstrated by the success of
spatial penalties in brain imaging. However the spatial extent of that
coupling should be small in brain imaging. Indeed, the typical scale used
to smooth fMRI data is of 6\,mm, for 3\,mm voxels. Similarly,
anatomical images, with voxels of 1\,mm, are often smoothed with a kernel
of 2\,mm. We use as a neighborhood $\mathcal{N}(i)$ of a voxel $i$ its 6
immediate neighbors. To let the behavior of a group be driven more by its
central voxel, we set the relative weights $\alpha$ of the 6 other voxels to
.7.

\paragraph{A FISTA solver}
The social-sparsity shrinkage $S$ is not the proximal 
operator of a known penalty. Yet, it has been shown to yield good
estimations in proximal optimization schemes \cite{kowalski2013social}.
We use a FISTA, an accelerated variant of ISTA, as for TV-$\ell_1$
\cite{dohmatob2014benchmarking,eickenberg2015total}. The brain-imaging
data fit appears via a loss $L: \B{w}\in \mathbb{R}^p \rightarrow
\mathbb{R}$, the gradient of which should be Lipschitz-continuous.
Typically, the logistic loss is used for classification problems and
the squared loss for regression. For completeness we detail the scheme in 
Algorithm~\ref{algo:fista}.

\begin{algorithm}[t]
    \SetKwInOut{Input}{input}\SetKwInOut{Output}{output}
    \Input{Initialization $\B{w}_{(0)} \in \mathbb{R}^{p}$,
	    ~penalization amount $\lambda$,
	   $L$-Lipschitz gradient of loss
	   $\nabla F$}
    $\B{v}_{(1)}\gets \B{w}_{(0)},\qquad$
    $k\gets 1,\qquad$
    $t_{(0)} \gets 0$\;
    \While{$\|\B{w}_{(k)} - \B{w}_{(k-1)}\|_\infty
		> \text{tol} \|\B{w}_{(k)}\|_\infty$}{
	    \(k \gets k + 1,\qquad\)
	    \(t_k\gets\frac{1}{2}\bigl(1 +
		\sqrt{1 + 4t_{(k-1)}^2}\bigr)\)\;
	    \(\B{w}_{(k)} \gets S\bigl(\B{v}_{(k)}
				- \frac{1}{L}\nabla F(\B{v}_{(k)}),\;
			\frac{\lambda}{L}\bigr)\)
	    \quad\text{$S$ given by (\ref{eq:social_shrinkage})}\;
	    \(\B{v}_{(k)}\gets \B{w}_{(k)} + 
		\frac{t_{(k-1)} - 1}{t_{(k)}}
		    (\B{w}_{(k)} - \B{w}_{(k - 1)})\)\;
    }
\caption{FISTA: social sparsity to estimate $\B{w}$}\label{algo:fista}
\end{algorithm}

\paragraph{Parameter selection}
We set the regularization parameter $\lambda$ by nested cross-validation,
using the same strategies as can be used in Graph-net or TV-$\ell_1$ to
speed up computation \cite{dohmatob2015speeding}. We do 8 folds. For each
fold, we scan the $\lambda$ parameter from large values to small values,
with warm start of the solver. As the model is similar to an $\ell_1$ penalty but with
additional constraints, values of $\lambda$ that lead pure $\ell_1$
models to be fully sparse will also give fully sparse social-sparsity
models. Hence we start our path at $\lambda_\text{max}$, the largest $\lambda$
giving non-empty $\ell_1$ model\footnote{
$\lambda_\text{max} = \|\B{X}^\mathsf{T} \B{y}\|_\infty$ for lasso, and
$\lambda_\text{max} = \frac{1}{n} \|\B{X}^\mathsf{T} \tilde{\B{y}}\|_\infty$ for $\ell_1$
logistic, where $\tilde{\B{y}}$ is the weighted output vector:
$\tilde{\B{y}}_i = \frac{n^+}{n}$ for samples in the positive class, and
$\tilde{\B{y}}_i = \frac{n^-}{n}$ for samples in the negative class, with
$n$ the number of samples, $n^+$ (resp.~$n^-$) the number in the positive 
(resp.~negative) class.
}. We visit 5 values on a logarithmic scale from $\lambda_\text{max}$ to
$\frac{1}{20} \lambda_\text{max}$.
As in Graph-net or TV-$\ell_1$ solvers \cite{dohmatob2015speeding},
we do early stopping of the optimizer on the left-out prediction error
during parameter selection.
The final coefficients are the average of the coefficient for the optimal
$\lambda$ on each of the 8 folds.

Finally, as it can be done with graph-net and TV-$\ell_1$, we use
univariate feature screening, retaining 20\% of the features. The
motivation for this screening is that sparse models are highly likely to
put the corresponding features to zero \cite{dohmatob2015speeding}.

\section{Experiments: Accuracy and run time} 

\paragraph*{Datasets} We study social sparsity on publicly-available
datasets for two applications: intra-subject brain-decoding from fMRI,
and inter-subject prediction from voxel-based morphometry (VBM).

For fMRI brain decoding, we use a standard visual-object recognition
dataset \cite{haxby2001}. 
We perform on all 5 subjects
intra-subject 2-class decoding of 14 pairs of stimuli of varying
difficulties (listed on Figure~\ref{fig:times_and_scores}). To measure
prediction accuracy, we perform cross-validation, leaving out 2 of the 12
sessions.

For VBM inter-subject prediction, we use the OASIS anatomical imaging
dataset \cite{marcus2007}. We predict gender and use a
cross-validation of random splits of 20\% of the subjects.


\paragraph*{Experimental settings}
On all classification tasks, we fit a model with a logistic loss and
measure prediction accuracy as well as wall time in ten iterations of
cross-validation.

We compare social sparsity to graph-net and TV-$\ell_1$, as implemented
in the Nilearn library, as well as the most commonly-used decoder, a
linear SVM with 20\% univariate feature selection. To fully replicate
common practice we use the default $C=1$ for the SVM, rather than
cross-validation. All models, including
social-sparsity, are implement in Python, using scikit-learn for the SVM
\cite{abraham2014machine}. The graph-net and TV-$\ell_1$ implementation 
use the same feature-selection and path strategies as our social-sparsity
solver to speed up computation.
An important detail for computation time is the
stopping criteria of the algorithms. We use the same for Graph-net,
TV-$\ell_1$, and social sparsity: a $10^{-4}$ cutoff on the relative
maximum change on the weights $\B{w}$ (see Alg.~\ref{algo:fista}).

\section{Results: Striking a good tradeoff}

\begin{figure}
\includegraphics[width=\linewidth]{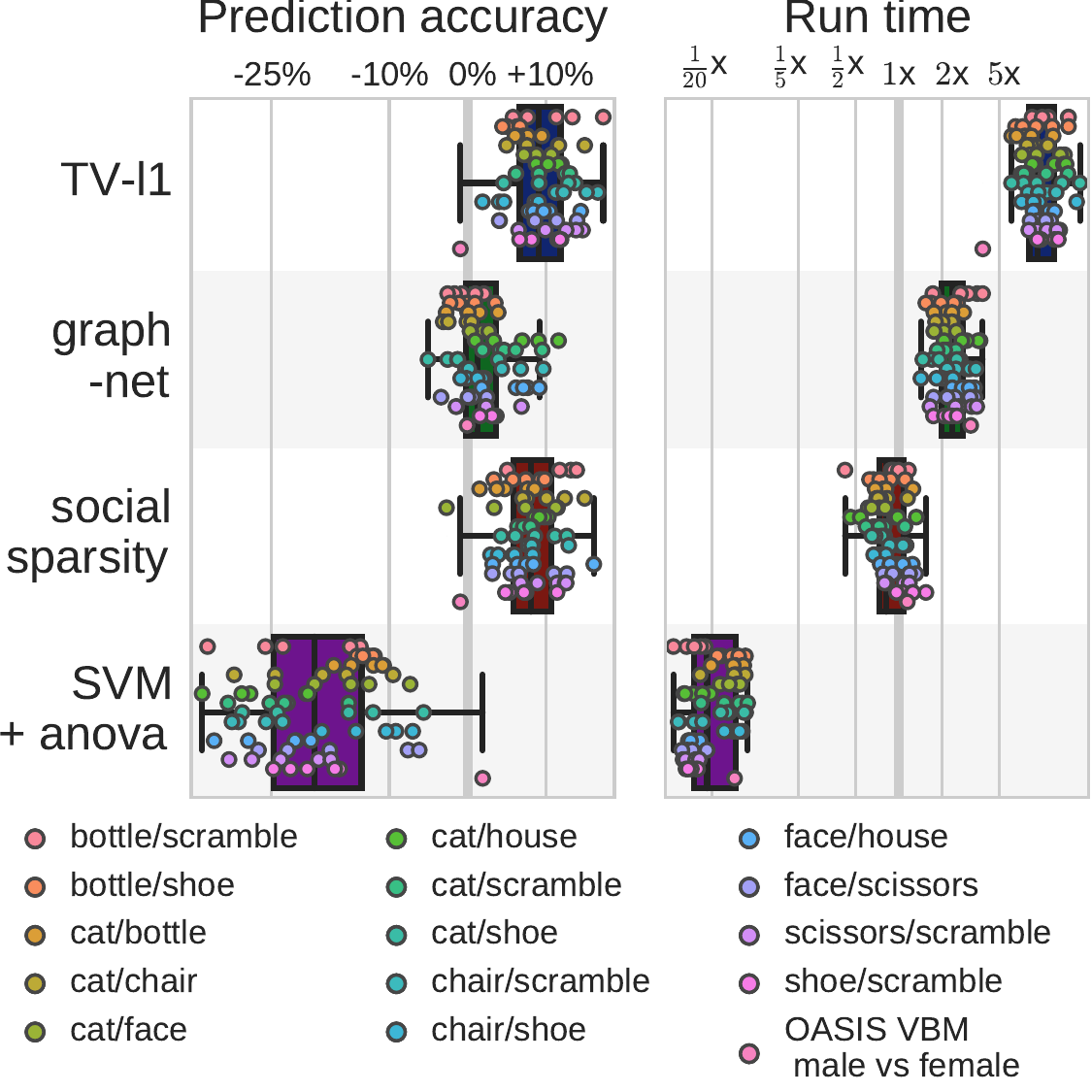}
\caption{\textbf{Prediction accuracy and computation time} for different
classification tasks from brain images: 14 on the Haxby dataset, and
gender prediction on OASIS. Values are displayed relative to the mean
over 4 classifiers: TV-$\ell_1$, graph-net, social sparsity,
and a linear SVM with 20\% univariate feature selection. Each subject of
the Haxby study gives one data point.
\label{fig:times_and_scores}}
\end{figure}

\begin{figure*}
\fboxsep0pt
\colorbox{black}{\begin{minipage}{\linewidth}
\hspace*{-.0215\linewidth}%
\includegraphics[height=.195\linewidth]{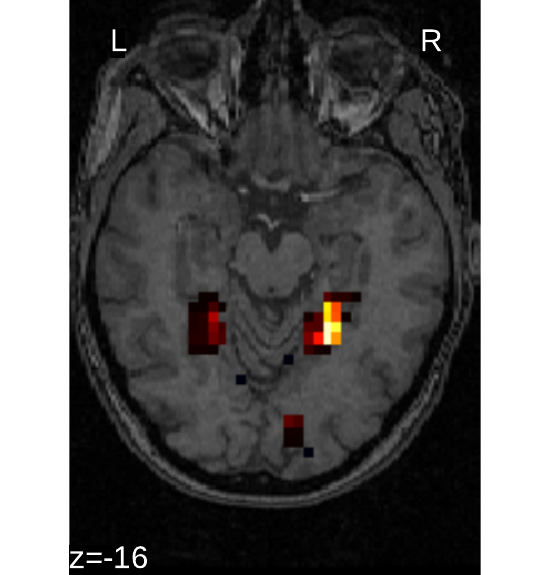}%
\hspace*{-.045\linewidth}%
\raisebox{.03\linewidth}{\smash{%
\includegraphics[trim=0cm 0cm 0cm 1cm, clip=true, height=.155\linewidth]{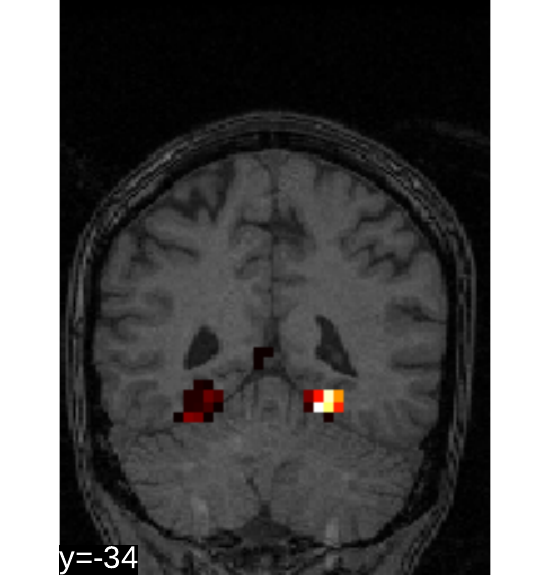}%
}}%
\hspace*{-.03\linewidth}%
\includegraphics[height=.195\linewidth]{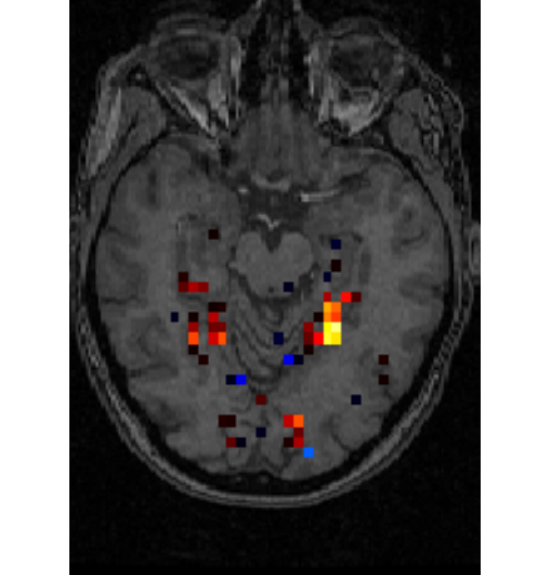}%
\hspace*{-.045\linewidth}%
\raisebox{.03\linewidth}{\smash{%
\includegraphics[trim=0cm 0cm 0cm 1cm, clip=true, height=.155\linewidth]{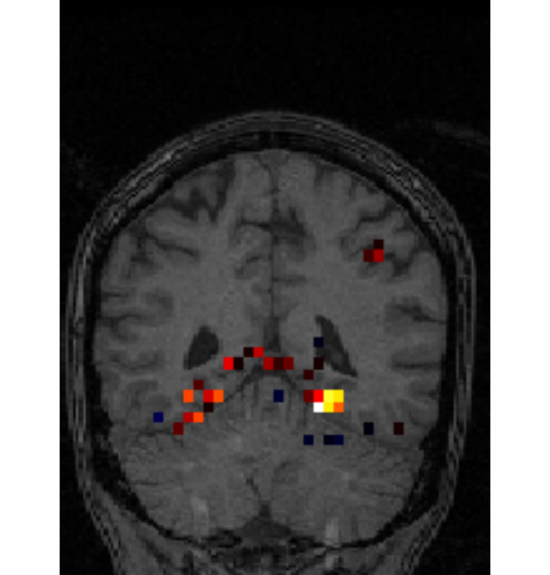}%
}}%
\hspace*{-.03\linewidth}%
\includegraphics[height=.195\linewidth]{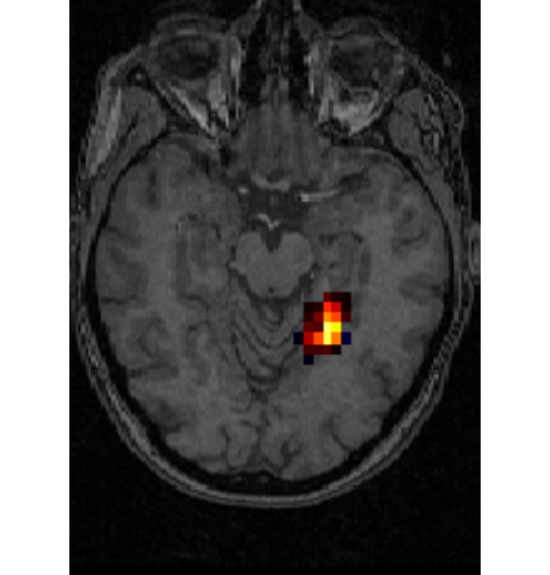}%
\hspace*{-.045\linewidth}%
\raisebox{.03\linewidth}{\smash{%
\includegraphics[trim=0cm 0cm 0cm 1cm, clip=true, height=.155\linewidth]{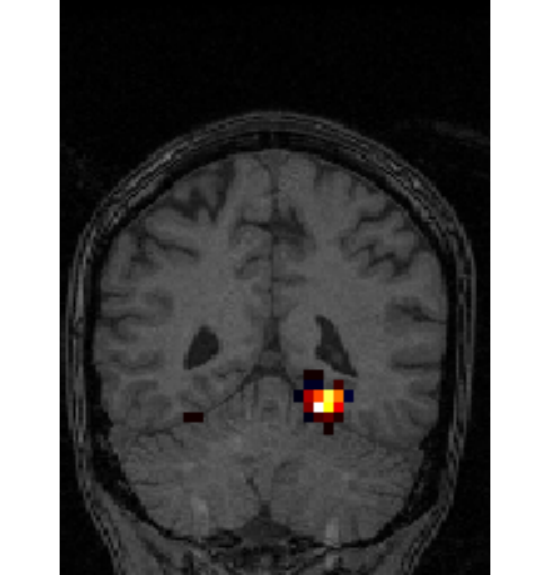}%
}}%
\hfill\raisebox{.16\linewidth}{\small\sffamily\color{white}\llap{face
vs~~\qquad\qquad}}%
\raisebox{.14\linewidth}{\small\sffamily\color{white}\llap{house~~\qquad\qquad}}%

\hspace*{-.0365\linewidth}%
\includegraphics[trim=0cm .5cm 0cm .5cm, clip=true, height=.187\linewidth]{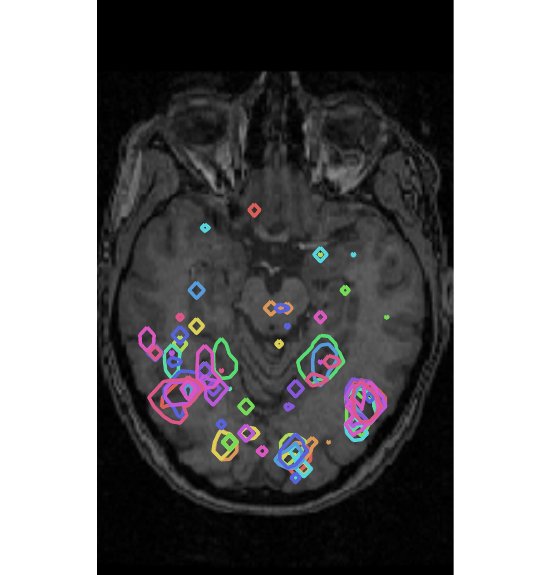}%
\hspace*{-.0758\linewidth}%
\includegraphics[trim=0cm .5cm 0cm .5cm, clip=true, height=.187\linewidth]{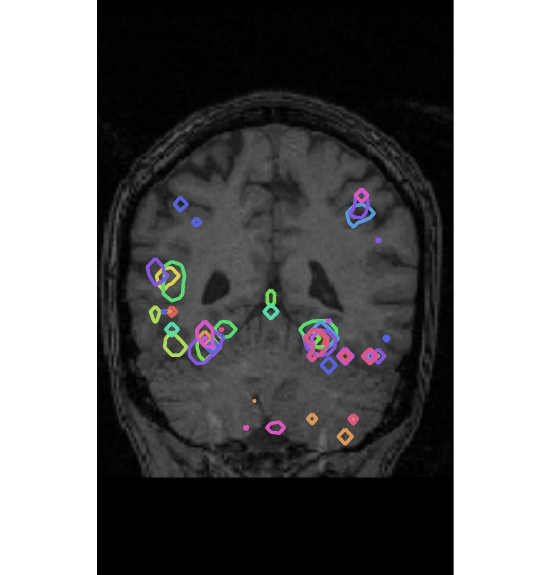}%
\llap{\smash{\raisebox{.18\linewidth}{\bfseries\sffamily\color{white}
TV-l1}\hspace*{.16\linewidth}}}%
\hspace*{-.0651\linewidth}%
\includegraphics[trim=0cm .5cm 0cm .5cm, clip=true, height=.187\linewidth]{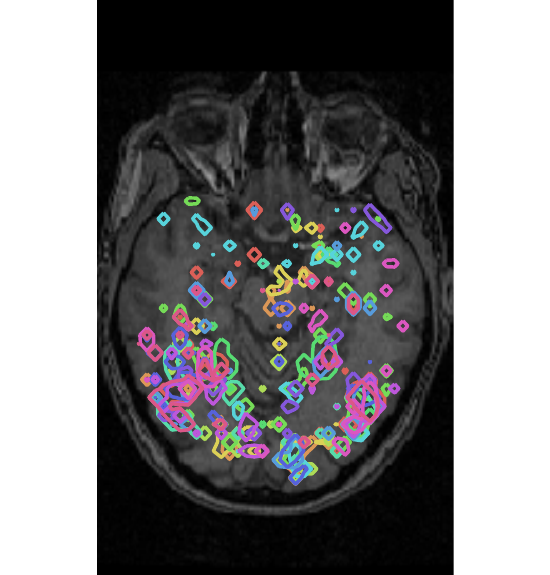}%
\hspace*{-.0758\linewidth}%
\includegraphics[trim=0cm .5cm 0cm .5cm, clip=true, height=.187\linewidth]{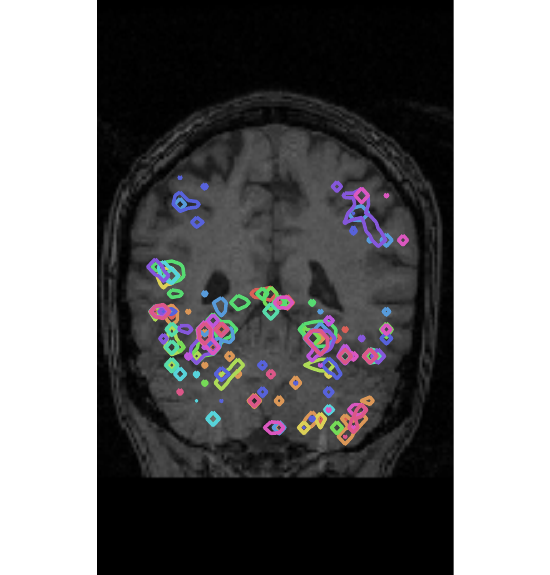}%
\llap{\smash{\raisebox{.18\linewidth}{\bfseries\sffamily\color{white}
Graph-net}\hspace*{.14\linewidth}}}%
\hspace*{-.0651\linewidth}%
\includegraphics[trim=0cm .5cm 0cm .5cm, clip=true, height=.187\linewidth]{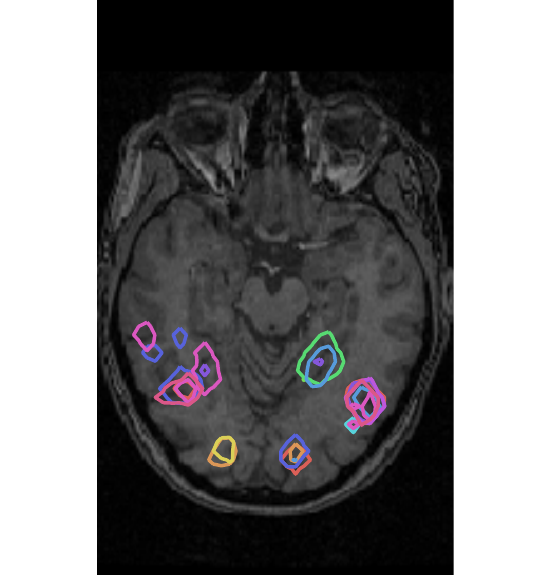}%
\hspace*{-.0758\linewidth}%
\includegraphics[trim=0cm .5cm 0cm .5cm, clip=true, height=.187\linewidth]{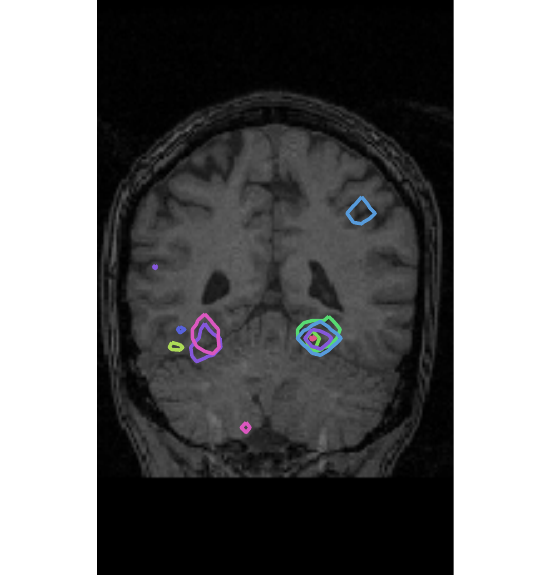}%
\llap{\smash{\raisebox{.18\linewidth}{\bfseries\sffamily\color{white}
Social sparsity}\hspace*{.1\linewidth}}}%
\hfill%
\hspace*{-.06\linewidth}%
\smash{%
\includegraphics[height=.22\linewidth]{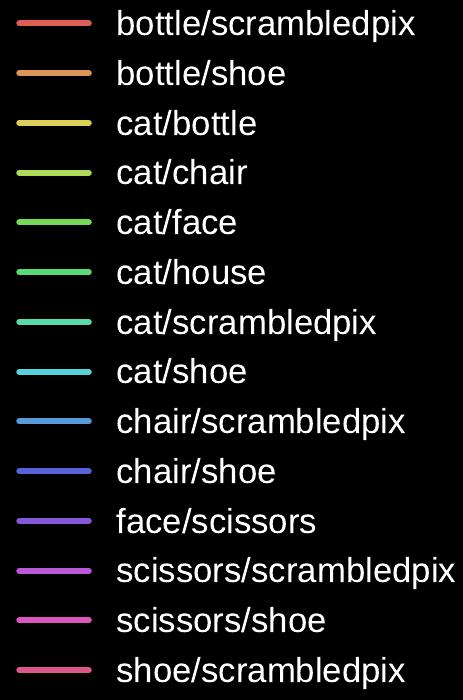}%
}%
\end{minipage}}
\caption{\textbf{Decoder maps for the object-classification task} --
  \textbf{Top}: weight maps for the face-versus-house task.  Overall,
  the maps segment the right and left parahippocampal place area
  (PPA), a well-known place-specific regions, although the left PPA is
  weak in TV-$\ell_1$, spotty in graph-net, and absent in social
  sparsity.  \textbf{Bottom}: outlines at $0.01$ of the other
  tasks. Beyond the PPA, several known functional regions stand out
  such as primary or secondary visual areas around the prestriate
  cortex as well as regions in the lateral occipital cortex,
  responding to structured objects \cite{grill2004human}.  Note that
  the graph-net outlines display scattered small regions even thought
  the value of the contours is chosen at $0.01$, well above numerical
  noise.
\label{fig:maps}}
\end{figure*}

Experiments outline a tradeoff between prediction accuracy and
computation time.
Fig.~\ref{fig:times_and_scores} displays the relative prediction accuracy and
run time. TV-$\ell_1$ predicts best on average over the various classification tasks. However, it is followed closely by social
sparsity which outperforms graph-net\footnote{All differences are
significant in a Wilcoxon rank test.}. The SVM performs much worse than
the spatial sparsity, aside from the VBM data where we find that all
models perform similarly.

In terms of run time, we find that graph-net is on average 4 times faster
than TV-$\ell_1$, but social sparsity is 3 times faster than graph-net.
The SVM-based decoder is 20 times faster than social sparsity, \emph{ie}
240 times faster than TV-$\ell_1$.

Finally, an important aspect of the brain decoders is whether they
segment well the brain regions that support the decoding. Such a question
is hard to validate, yet there is evidence that TV-$\ell_1$ is a good
approach \cite{gramfort2013}. Fig.~\ref{fig:maps} displays the decoder
maps for the object-recognition tasks. For these tasks, we expect
prediction to be driven by the functional areas of the visual
cortex \cite{grill2004human}. Indeed, the maps outline regions in known
visual areas. The graph-net maps are much more scattered and less
structured than the others. Conversely, the social sparsity maps are
sparser and outline a smaller number of clusters.

\section{Conclusion: Be social}

Brain decoders benefit strongly from spatial sparsity that helps them
narrow on regions important for prediction. Total variation is a powerful
and principled solution, but it comes with a hefty computational
cost. Social sparsity can be used to introduce penalties on local
neighborhoods in the image. It is more heuristic, as it does not
minimize a known convex cost. We find empirically that it performs very
slightly less well that TV-$\ell_1$ in terms of prediction accuracy, but is more than ten
times faster. The corresponding decoder maps are more sparse and focus on
a smaller number of regions than TV-$\ell_1$. Social sparsity outperforms
graph-net, the faster contender to TV-$\ell_1$, on speed, accuracy, and
interpretability. We have found that it strikes a very interesting
balance for brain decoding: much faster and almost as good for prediction
as TV-$\ell_1$. A full social-sparsity model-fit with hyper-parameter
selection takes only 20 times longer than a simple SVM with default
hyper-parameters. With social-sparsity, spatially-structured brain
decoders are fast: typically 30\,s
with parameter selection on a 2\,GHz i7 CPU.

\paragraph*{Acknowledgment}
{\small
OASIS was supported by grants P50 AG05681, P01 AG03991, R01 AG021910, P50 MH071616, U24 RR021382, R01 MH56584. 
The authors acknowledge funding from the
EU FP7/2007-2013 under grant agreement 604102 (HBP).
}
%
%
\bibliographystyle{IEEEtran} 
\bibliography{biblio} 
\end{document}